\documentclass[10pt,twocolumn,letterpaper]{article}

\usepackage{cvpr}
\usepackage{times}
\usepackage{epsfig}
\usepackage{graphicx}
\usepackage{amsmath}
\usepackage{amssymb}
\usepackage{mathptmx}
\usepackage{bm}
\usepackage{color}
\usepackage{multirow}
\usepackage{subcaption}
\newcommand{\hwchange}[1]{\textcolor{black}{#1}}
\newcommand{\mr}[1]{\multirow{2}{*}{#1}}

\usepackage{wrapfig}
\usepackage[symbol]{footmisc}

\usepackage[breaklinks=true,bookmarks=false]{hyperref}

\cvprfinalcopy 

\def\cvprPaperID{5} 

\begin{document}

\title{ANTNets: Mobile Convolutional Neural Networks for Resource Efficient Image Classification}

\author{Yunyang Xiong\thanks{Work done at Amazon Lab126.}\\\
University of Wisconsin-Madison\\
{\tt\small yxiong43@wisc.edu}
\and
Hyunwoo J. Ki$\text{m}^{\ast}$ \\
Korea University\\
{\tt\small hyunwoojkim@korea.ac.kr}
\and
Varsha Hedau\\
Amazon Inc.\\
{\tt\small hedauv@amazon.com}
}
\newcommand{\hjk}[1]{\textcolor{black}{#1} }
\maketitle

\begin{abstract}
Deep convolutional neural networks have achieved remarkable success in
computer vision. However, deep neural networks require large computing resources
 to achieve high performance. Although depthwise separable
convolution can be an efficient module to approximate a standard convolution, 
it often leads to reduced representational power of networks. In this paper, under
budget constraints such as computational cost (MAdds) and the parameter count,
we propose a novel basic architectural block, ANTBlock. It boosts the
representational power by modeling, in a high dimensional space, interdependency
of channels between a depthwise convolution layer and a projection layer in the
ANTBlocks.
Our experiments show that ANTNet built by a sequence of ANTBlocks,
 consistently outperforms state-of-the-art
low-cost mobile convolutional neural networks across multiple datasets. On
CIFAR100, our model achieves $75.7\%$ top-1 accuracy, which is $1.5\%$ higher
than MobileNetV2 with 8.3\% fewer parameters and 19.6\% less computational cost.
On ImageNet, our model achieves $72.8\%$ top-1 accuracy, which is $0.8\%$
improvement, with $157.7ms$ ($20\%$ faster) on iPhone 5s over MobileNetV2.
\end{abstract}

\section{Introduction}
Deep neural networks have emerged as state-of-the-art solutions for various
tasks in computer vision, machine learning, and natural language processing.
Recent research in deep learning mainly focuses on deeper and heavier models to
achieve superhuman accuracy with a tremendous number of parameters. Inception
\cite{ioffe2015batch}, ResNets \cite{he2016deep}, HighwayNets
\cite{srivastava2015training}, and DenseNet \cite{huang2017densely} are popular
architectures in this direction and has been shown to be effective in a variety of tasks.
However, in many real-world applications, due to limited computing resources and
short latency requirements, more efficient recognition systems are often required,
for example, in mobile phones, robots, and smart appliances that require the on-device
intelligence systems.

For the last few years, small and efficient neural networks have enabled
the deployment of models on computationally limited hardware for a wide range of
applications. One stream of such efforts is to substitute existing layers with
more efficient layers. Since in a vision system, convolutional neural networks
(CNNs) are the most popular base feature extraction networks and the main
computational burden is convolutional layers, faster convolution layers are
crucial. The standard convolution layer performs convolution using all the input
channels for one output channel. So, the number of filters and calculations
increase as the number of input channels grows. Instead, \textit{group
convolution} involves only each group of input channels resulting in a smaller
number of filters (and calculations) reduced by a factor of the number of
groups. Group convolution has been used in multiple architectures. AlexNet
\cite{nips2012_4824} uses group convolution to train models on GPUs with limited
memory. Later, ResNeXt \cite{xie2017aggregated} utilizes group convolution to
achieve better performance and \cite{ioannou2017deep} proposed more complex
group convolution with hierarchical arrangements. One extreme of group
convolution is \textit{depthwise separable convolutions} introduced in
\cite{sifre2014rigid}. Each group involves only one input channel and
convolution filter. Since then, the trick has been adopted in other
architectures such as Inception \cite{ioffe2015batch}, Flattened Networks
\cite{jin2014flattened} and Xception \cite{chollet2017xception}. Recently, the
depthwise separable convolution has been adopted by a compact architecture
specifically designed for mobile devices. MobileNetV1
\cite{howard2017mobilenets}, and MobileNetV2 \cite{sandler2018mobilenetv2}
achieved significant improvement with respect to inference time (latency) on
mobile devices.

Efficient convolutional layers are preferable but accuracy degradation is
inevitable. To fill the performance gap induced by approximate convoltuion,
recent network enhancement techniques can be used as long as the additional cost
is negligible.
There have been multiple attempts to boost the representational power of models
with negligible additional cost. Attentional neural networks have been proven
that it is a general module and improve performance by suppressing irrelevant
information and focus on informative parts of data. Temporal/spatial attention
has been studied in the literature but, arguably, they come with a significant cost.
However, channel attention can be implemented in a much more efficient way. For
instance, the ``Squeeze-and-Excitation'' (SE) block proposed in SENet
\cite{hu2018senet} allows selective reweighting channels based on global
information from each channel. This improves a variety of architectures with a
minor computational cost. A channel shuffle operation also boosts the performance
or mitigates degradation of group convolution as shown in ShuffleNet
\cite{Zhang_2018_CVPR,ma2018shufflenet} and two-stage convolution
\cite{xie2018interleaved}. This line of efforts motivate our work to develop
an efficient and powerful architecture.

\hjk{ Our \textbf{contributions}: \textbf{(i)} We propose a new efficient and
powerful architectural block, ANTBlock, that dynamically utilizes channel
relationships;
\textbf{(ii)} we show that a naive adaptation of channel
attention (e.g., SE) does not improve the representational power of depthwise
convolutional layers and propose an optimal
configuration that maximizes the number of channels and has \textit{full channel receptive fields};
\textbf{(iii)} using group convolution we make the ANTBlock more efficient w.r.t.
parameter counts and computational costs without significant performance
loss and extend it to an ensemble block; \textbf{(iv)} ANTBlock is simple to
implement in widely-used deep learning frameworks and
outperforms the state-of-the-art lightweight CNNs. ANTNet achieves $0.8\%$
improvement over MobileNetV2 on the ImageNet
\cite{russakovsky2015imagenet} with $6\%$ fewer parameters and $10\%$ fewer
multiply-adds (MAdds) resulting in 20\% faster inference time on a mobile phone. }
%


\section{Related Work} \label{sec:relatedwork}
%

The efficiency of neural networks becomes an important topic as networks get
larger and deeper. Inception module was utilized in GoogLeNet
\cite{szegedy2015going} to obtain high performance with a drastically reduced
number of parameters by using small convolutions. An efficient bottleneck structure was
designed to construct ResNet to achieve high performance. Further, the large
demand for on-device applications encourages studies on resource-efficient
models with minimal latency and memory usage. To this end,
\cite{he2015convolutional} studied module designs with the trade-off between
multiple factors such as depth, width, filter size, pooling layer and so on.

Group convolution is a straightforward and effective technique to save
computations while maintaining accuracy. It was introduced with AlexNet
\cite{krizhevsky2012imagenet} as a workaround for small GPUs. Later DeepRoots
\cite{ioannou2017deep} and ResNeXt \cite{xie2017aggregated} adopted group
convolutions to improve models. Depthwise separable convolution is an extreme
case of group convolution that performs convolution each channel
separately. It was first introduced in \cite{sifre2014rigid} and Xception
\cite{chollet2017xception} integrates the idea into the Inception and
CondenseNet \cite{Huang_2018_CVPR} does so for DenseNet.
For mobile platforms, MobileNetV1 \cite{howard2017mobilenets}, and
MobileNetV2 \cite{sandler2018mobilenetv2} used depthwise convolutions with some
hyperparameters to control the size of models.

Channel relationship is a relatively underexplored source of the performance
boost. It is a promising direction since it usually requires a small additional
cost. ShuffleNets \cite{Zhang_2018_CVPR,ma2018shufflenet} shuffle channels
within two-stage group convolution and can be efficiently implemented by
``random sparse convolution'' layer
\cite{changpinyo2017power,zhang2017interleaved}. Apart from random sparse
channel grouping, Squeeze-and-Excitation Networks (SENet) \cite{hu2018senet}
studies a dynamic channel reweighting scheme to boost model capacity at a small
cost. The success of channel grouping and channel manipuliation
motivates our work.


\section{Model Architecture: ANTNet} \label{sec:arch}
The goal of this work is to design a basic low cost architecture block, which
can be used to build efficient Convolutional Neural Networks for mobile devices
with budget constraints. The budget of a model varies depending on
implementations and hardware quantities for real-world applications. To have a
general and fair comparison, in the literature \cite{han2015learning}, the
budget (or complexity) of a model is measured by the number of computations,
e.g., multiply-adds (MAdds) or floating point operations (FLOPs), and the number of model parameters. Our goal is to build a more accurate CNN, ANTNet (\textbf{A}ttention \textbf{N}es\textbf{T}ed \textbf{Net}work),  with fewer \textit{MAdds} and \textit{Params} by stacking our novel basic blocks. ANTNet utilizes depthwise separable convolution and channel attention. Before introducing our ANTBlock, we breifly discuss depthwise seprable convolution and its variations with computation budgets (i.e., MAdds, and Params).

\textbf{Depthwise Separable Convolution} is
proven to be a effective module to build efficient neural network architectures. It
approximates a standard convolution operation with two separate convolutions:
 depthwise convolution and pointwise convolution.
The most common depthwise separable convolution \cite{howard2017mobilenets}
consists of two layers: a $3\times3$ depthwise convolution that filters the data
and $1\times1$ point-wise convolution that combines the outputs of depthwise
convolution. Consider that the input and output of the depthwise separable
convolution are three dimensional feature maps of size $H_1\times W_1 \times
C_1$, $H_2\times W_2 \times C_2$, where $H_i$, $W_i$, and $C_i$ denote height, width,
and the number of channels of the feature map and $i$ indicates input $(i=1)$ and output $(i=2)$.
 $H_1, H_2$ are the height, $W_1, W_2$
 are width, and $C_1, C_2$ are the number of channels of the input and output
 feature map.
For a convolution kernel size $K\times K$, 
The total number of MAdds for depthwise separable convolution is
\begin{equation}
(K \times K + C_2) \times C_1 \times H_2 \times W_2.
\end{equation}
Compared to the standard convolution, it reduces almost $K \times K$ times
computational cost. However, it often leads to reduced representational power.

\begin{table*}[!h]
	\centering
	{\footnotesize \begin{tabular} {|c|c|c|c|}
			\hline
			Layer & Input & Operator & Output\\
			\hline
			(a) Expansion layer & $H \times W \times C_1$ & 1x1 conv2d, ReLU6 & $H \times W \times (C_1 \times t)$ \\
			\hline
			(b) Depthwise layer & $H \times W \times (C_1 \times t)$ & 3x3 dwise stride = $s$, ReLU6
			& $H/s\times W/s\times (C_1 \times t)$ \\
			\hline
			(c) Channel attention layer & $H \times W \times (C_1 \times t)$ & Global pooling, FC(2), Sigmoid
			& $H/s\times W/s\times (C_1 \times t)$ \\
			\hline
			(d) Group-wise projection layer & $H/s\times W/s\times (C_1 \times t)$ & linear 1x1 gconv2d group = $g$ & $H/s\times W/s\times C_2$ \\
			\hline
		\end{tabular}
	}
	\caption{\label{tab:antblock} \footnotesize The structure of ANTBlock that it
		transforms from $C_1$ to $C_2$ channels with expansion factor $t$, group $g$
		with stride $s$.  }
	\vspace{10pt}
\end{table*}

\textbf{Inverted Residual Block.}
One quick fix for the reduced representational power is to increase the input
channels for depthwise separable convolution by adding an expansion layer before
depthwise convolution in inverted residual bottleneck blocks of MobileNetV2
\cite{sandler2018mobilenetv2}. The expansion operation expands the number of
input channels to $t$ times by $1\times 1$ point-wise convolution.
The inverted residual block has three different types of layers, expansion layer, depthwise layer and projection layer. The projection layer takes the largest portion of computation and more parameters
than others when the number of input/output channels $C$ is larger than kernel parameters $K \times K$.
 Based on our observation on MAdds and parameters of inverted residual block, we will develop
a more accurate and efficient block by saving computations on the projection
layer with cheaper operations and allocating more resource on the depthwise
layer with a small additional cost.


\subsection{Designing Efficient Blocks: ANTBlock}
\begin{figure}[!b]
	\begin{subfigure}[] {0.49 \linewidth}
		\centering
		\includegraphics[clip, trim=2.25cm 12.5cm 16cm 2cm]{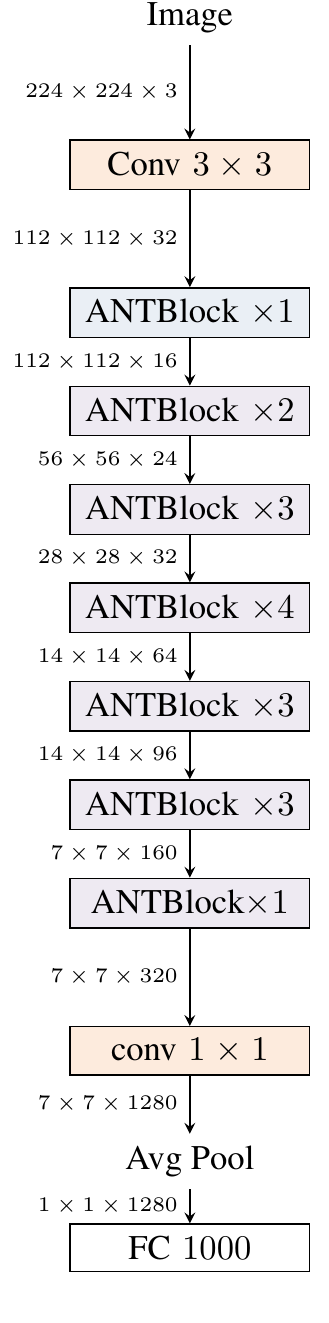}
		\caption{ANTNet}
	\end{subfigure}
	\begin{subfigure}[] {0.49 \linewidth}
		\centering
		\includegraphics[clip, trim=2.25cm 13cm 15cm 1.5cm]{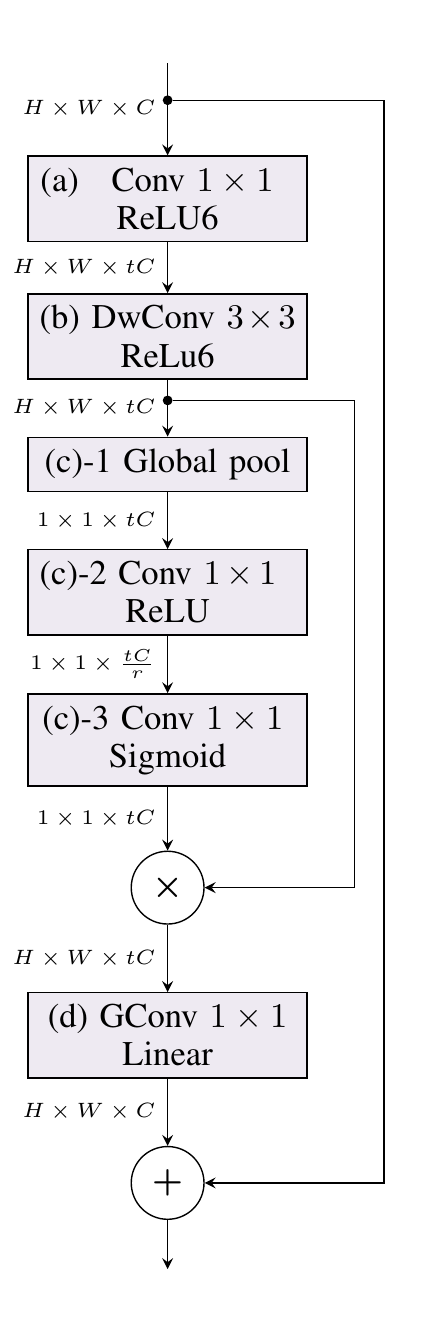}
		\caption{ANTBlock}
	\end{subfigure}
	\caption{\label{fig:antnetblock}\footnotesize ANTNet architecture for ImageNet.
		$H\times W\times C$ is the dimension of the tensor, $t$ is the expansion factor
		of channels, $r$ is the reduction ratio for channel attention.
		Symbol $\bm{\oplus}$ denotes the element-wise addition and symbol
		$\bm{\otimes}$ denotes the channel-wise multiplication. ANTBlock $\times
		1/2/3/4$ means the number of repeated layers within ANTBlock. Note that If the
		output resolution differs from the input resolution, only the stride of the
		first layer within ANTBlock is $=2$ and residual connection of the block is
		skipped. DwConv stands for depthwise convolution and GConv stands for group
		convolution.  (a): ANTNet model is shown in Table \ref{tab:antnet} with more
		details; (b) is the structure of corresponding ANTBlock for building ANTNet and
		it is shown in Table \ref{tab:antblock}. }
\end{figure}

In this section, we introduce our ANTBlock with detailed
	discussion. The ANTBlock presented in Fig. \ref{fig:antnetblock}(b)
	is a residual block and can be written as

\begin{equation}
\label{eq:invres}
\tilde{\bm{x}} = \bm{x} + F(\bm{x}).
\end{equation}
When the dimension of the input to the block is not the same as
	the output, i.e., $(\text{dim}(\bm{x}) \neq \text{dim}(\tilde{\bm{x}}))$, we
	simply skip the residual connection as MobileNet V2. For simplicity, let us
	focus on the equation \ref{eq:invres}. ANTBlock is motivated by the Inverted
	Residual Block in MobileNet V2 and it can be factored into two parts:
	mapping $G(\cdot)$ from the input space to the high dimensional depthwise convolution space, $\mathbf{R}^{H \times W \times tC}$,
	and  projection $H(\cdot)$ to the input space, $\mathbf{R}^{H \times W \times C}$.
	Then, Eq. \ref{eq:invres} can be written as
\begin{equation}
\label{eq:facF}
F(\bm{x}) = H(G(\bm{x})).
\end{equation}
\hwchange{In ANTBlock, $G(\cdot)$ consists of one expansion layer and one
	depthwise convolutional layer.
	$H(\cdot)$ is a projection layer.
	Now, our block can be rewritten as}
\begin{equation}
\label{eq:invF}
\tilde{\bm{x}} = \bm{x} + H(G(\bm{x}))
\end{equation}
\hwchange{This construction can be further improved by the attention mechanism.
	In \cite{long2015fully}, the models equipped with attention mask $M(\bm{x})$ show
	significant improvement on segmentation.}
We apply this similar idea to the output of depthwise
layer for boosting feature representation. In this case, channel attention is  used to improve representational
power without a significant increase in computational cost and parameters. With channel attention, we can write our ANTBlock (see Fig. \ref{fig:antnetblock} (b)) as,
\begin{align}
\label{eq:AttInvF}
\tilde{\bm{x}}_c = \bm{x}_c + H_c(M(G(\bm{x}))*G(\bm{x})), \forall c \in \{1, \cdots, C\},
\end{align}
\hwchange{where $*$ stands for element-wise product, $\bm{x}$ is the input feature, corresponding to the input of ANTBlock in Fig. \ref{fig:antnetblock} (b). $G(\bm{x})$ denotes the output of depthwise convolution layer (b) of ANTBlock. $M(G(\bm{x}))$ denotes the attention mask for $G(\bm{x})$, represented by (c)-1, (c)-2 and (c)-3 in Fig. \ref{fig:antnetblock} (b). $c$ denotes an output channel of ANTBlock, $H_c$ is the projection for each output channel, corresponding to (d) Fig. \ref{fig:antnetblock} (b) with group convolution (group $g$ = 1), which means the output of ANTBlock is using all features from the output of attention maps $M(G(\bm{x}))*G(\bm{x})$.}

\begin{figure}[!b]
	\centering
	\includegraphics[clip, trim=2.25cm 12.5cm 11cm 2cm]{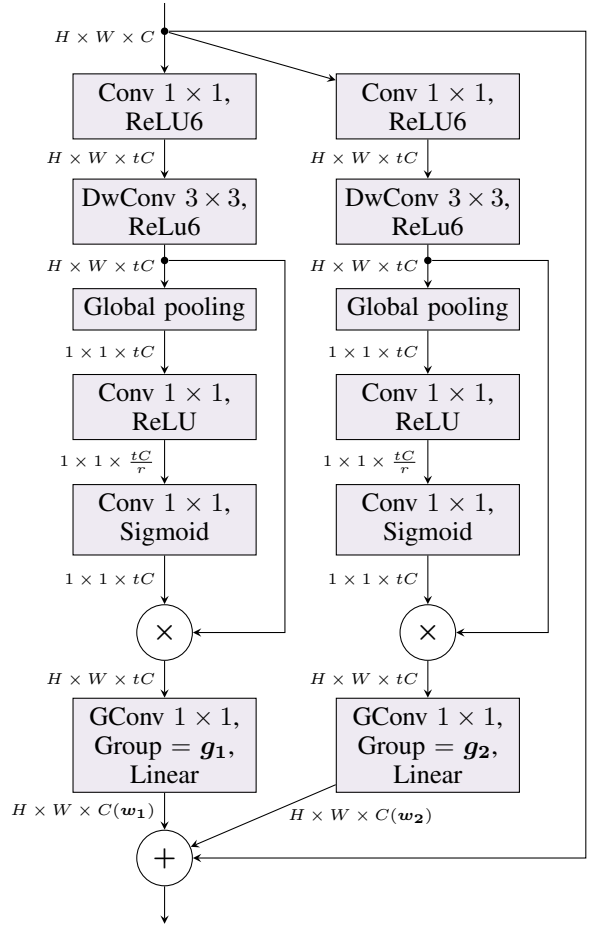}
	\vspace{-30pt}
	\caption{\label{fig:eantblock}\footnotesize The structure of e-ANTBlock for building e-ANTNet. Two types of ANTBlock are used for constructing e-ANTBlock and the weight parameters $w_1$ and $w_2$ of each e-ANTBlock are learned end-to-end for training e-ANTNet built on a sequence of e-ANTBlock. The e-ANTNet architecture is similar to ANTNet except that the ANTBlocks of ANTNet are replaced with e-ANTBlocks. }
\end{figure}
	As discussed before, the parameters and MAdds of depthwise convolutional layer
	kernels are usually fewer than expansion layer and projection layer. We use a group convolutional layer forward more efficient projection
	saving parameters and MAdds by a factor of groups.
	Group convolution first has been adopted in
	\cite{krizhevsky2012imagenet} to use multiple GPUs for distributed convolution
	computation. It reduces computational cost and the number of parameters while
	still achieving high representational power.
	\cite{zhang2017clcnet} proposed \textit{channel local convolution} (CLC), in which an
	output channel can depend on an arbitrary subset of the input channels.
	It is a multi-stage group convolution with a nice property so-call \textit{full channel receptive field}
	(FCRF).
	They found that in order to achieve high accuracy  every output channel of (CLC)
	should cover all the input channels.
	In our case, channel attention uses all the input channels of the depthwise
	convolution layer. So any group convolution for the projection layer satisfies FCRF
	condition and our ANTBlock  becomes a CLC block.
	With a group convolution layer for the projection layer $H(\cdot)$,
	our ANTBlock can be written as
\begin{align}
\label{eq:AllInvF}
\tilde{\bm{x}}_c = \bm{x}_c + H_c(& M_{t +1 \cdots t +
	\frac{C'}{g} }(G(\bm{x}))*G_{t + 1, \cdots t + \frac{C'}{g}}(\bm{x})), \\
& \forall c \in \{1, \cdots, C\}, \nonumber
\end{align}
\hwchange{where $g$ is the number of group convolution, $C'$ denotes the output channels of $G(\bm{x})$, $t = c \pmod g\times
	\frac{C'}{g}$, $M_{t +1 \cdots t +
		\frac{C'}{g} }(G(\bm{x}))*G_{t + 1, \cdots t + \frac{C'}{g}}(\bm{x})$ denotes the $\frac{C'}{g}$ feature
	maps associated with each output channel $c$ of ANTBlock. $H_c$ is the projection for each output channel with group convolution (group $g$), corresponding to (d) Fig. \ref{fig:antnetblock} (b).}
\subsection{Ensemble ANTBlocks: e-ANTBlock}
\hjk{The proposed block (ANTBlock) can be extended further to an ensemble block,
denoted by e-ANTBlock. To construct more powerful networks, we can ensemble (or weighted aggregate)
 different types of ANTBlocks (e.g., different group). e-ANTBlock can be written as
\begin{equation}
	\label{eq:e-ANTBlock}
	\tilde{\bm{x}} = \bm{x} + \sum_{j=1}^{m}w_jF_j(\bm{x})
\end{equation}
where $m$ is the number of different ANTBlocks,
$F_j$ denotes an ANTBlock with a group convolutional layer for projection,
$w_j$ is a weight of an ANTBlock. The weights $\{w_j\}_{j=1}^m$ are outputs of a softmax function
written as
\begin{equation}
w_j = \frac{e^{\lambda_j}}{\sum_{j = 1}e^{\lambda_j}}, 1\leq j \leq m.
\end{equation}
so that $\sum_{j = 1}^{m}w_j = 1$ and $\forall w_j \in [0, 1]$.
$\{ \lambda_j \}_{j=1}^m$ are parameters of e-ANTBlock trained by backpropagation. During
standard training, these parameters of e-ANTBlock will be learned end-to-end. In our experiments, $j \in \{1,2\}$ is used. The structure of e-ANTBlock with $m = 2$ can be seen in Fig. \ref{fig:eantblock}
}


\subsection{ANTNet}
\hwchange{ANTNet (\textbf{A}ttention \textbf{N}es\textbf{T}ed \textbf{Net}work) is a new
efficient convolutional neural network architecture constructed by a sequence of
ANTBlocks. The architecture of the network is similar to MobileNetV2, but all
the inverted residual blocks are replaced by ANTBlocks and one may use a
different number of ANTBlocks depending on the target accuracy.}

\hwchange{Now, we describe our architecture in detail. The basic building block
is ANTBlock, which has an expansion layer, a depthwise convolutional layer, a
channel-attention layer, and a group-wise projection layer with residual
connections. The detailed structure of ANTBlock is shown is Table
\ref{tab:antblock} and Figure \ref{fig:antnetblock}(b).}

\begin{table*}[!htbp]
	\centering
		{\footnotesize \begin{tabular} {|c|c|c|c|c|c|c|c|c|}
			\hline
			\mr{Name}&\mr{Input}&\mr{Operator}& \mr{Expansion ($t$)} &Reduction & Output  & Repetition ($n_i$) & \mr{Stride ($s$)} & \mr {Group ($g$)}\\
			      &          &          &                 &  Ratio ($r$)     & Channels ($C_{\text{out}}$) &                  &              &             \\
			\hline
			conv0 &$224 \times 224 \times 3$ & conv2d & - & - & 32 & 1 & 2 & - \\
			\hline
			ant1 &$112 \times 112 \times 32$ & ANTBlock & 1 & 8 & 16 & 1 & 1 & 1 \\
			\hline
			ant2&$112 \times 112 \times 16$ & ANTBlock & 6 & 8 & 24 & $2$ & 2 & 2 \\
			\hline
			ant3&$56 \times 56 \times 24$ & ANTBlock & 6 & 12 & 32 & $3$ & 2 & 2 \\
			\hline
			ant4&$28 \times 28 \times 32$ & ANTBlock & 6 & 16 & 64 & $4$ & 2 & 2 \\
			\hline
			ant5&$14 \times 14 \times 64$ & ANTBlock & 6 & 24 & 96 & 3 & 1 & 2 \\
			\hline
			ant6&$14 \times 14 \times 96$ & ANTBlock & 6 & 32 & 160 & $3$ & 2 & 2 \\
			\hline
			ant7&$7 \times 7 \times 160$ & ANTBlock & 6 & 64 & 320 & $1$ & 1 & 2 \\
			\hline
			conv8&$7 \times 7 \times 320$ & conv2d 1x1 & - & - & 1280 & 1 & 1 & - \\
			\hline
			pool9&$7 \times 7 \times 1280$ & avgpool 7x7 & - & - & 1280 & 1 & 1 & - \\
			\hline
			fc10&$1 \times 1 \times 1280$ & FC & - & - & n & - & - & - \\
			\hline
		\end{tabular}
	}
\caption{\label{tab:antnet} \footnotesize The architecture of ANTNet($g=2$). Each line
gives a sequence of $1$ or more identical (modulo stride) layers with repetition
times. All layers in the same module or sequence have the same number of output
channels. The stride $2$ is applied to the only first block in each layer.
ANTNet($g=1$) has the same parameters as above but $g$ is always $1$.}
\vspace{10pt}
\end{table*}

\hwchange{A channel attention block (c) in Fig.\ref{fig:antnetblock}-(b) introduces additional parameters and MAdds compared to
the Inverted Residual Blocks.
Consider our ANTNet which has $N=7$ group of repeated blocks as shown in Table \ref{tab:antnet}.
Each group has $n_i$ ANTBlocks.
Given reduction ratio $r_i$,
the increase in computational cost can be written as}
\begin{align}\label{eq:cparams}
	2 * \sum_{i=1}^{N} n_i \cdot \frac{{\left (C_i^{'}\right )}^2}{r_i},
\end{align}
\hwchange{where $C_i^{'}$ is the number of output channels from the depthwise convolutional layer.
The equation \ref{eq:cparams} shows that when the dimension of output channels
increases, the number of additional parameters and MAdds will increase. Also,
\cite{hu2017squeeze} demonstrates that the channel attention is prone to be saturated
at later layers and saturation also appear in our experiments. Therefore, the reduction ratio $r_i$ can be optimized
for each repeated blocks and our configuration is shown in Table \ref{tab:antnet}. Later layers have less degree of freedom in terms of channel reweighting.}

The group-wise projection layer in the ANTBlock reduces the number of parameters
and MAdds. The group parameters $g$ needs to be determined for every ANTBlock
for building ANTNet. It is the trade-off between efficiency and model accuracy.
The overall design of ANTNet is shown in Table \ref{tab:antnet}. We set the
expansion rate $t=1$ and $g=1$ for the first repeated blocks (ant0). In
others, we use a constant expansion rate and a group throughout the network.

\hjk{When more budgets (e.g., \textit{MAdds} and \textit{Params}) are allowed,
e-ANTBlock can be used as a basic block to build a more powerful network.
e-ANTNet achieves the highest accuracy as shown in Table \ref{tab:cifarperf}.}

\section{Experiments} \label{sec:expr}
We evaluate the computational efficiency and accuracy of ANTNet and compare
it with state-of-the-art mobile models with favorable classification accuracy.
The computational efficiency is measured theoretically by MAdds, and Params, and
empirically by CPU Latency and model size on a mobile platform (iPhone 5s). For
accuracy of models, we evaluate the image classification accuracy on CIFAR100
dataset \cite{krizhevsky2009learning} and ImageNet dataset (ILSVRC 2012 image
classification) \cite{russakovsky2015imagenet}. For ImageNet, we follow the
prior work and use the validation dataset as a test set.

ANTNet is implemented using PyTorch. We use built-in 1$\times$1 convolution and
group convolution implementation for channel attention, and projection layers.
Our ANTBlock is easy to reproduce in any deep learning frameworks such as
Caffe, TensorFlow, and MXNet using built-in layers as long as $1\times$1 standard
convolution and group convolution are available.

SGD optimizer was used in our experiments for model training. The
momentum of SGD optimizer is set to $0.9$ and the \textit{nesterov} momentum is used.
We use a multistep learning rate schedule with initial learning rate $0.01$ and
multiplicative factor of learning rate decay $\gamma=0.1$ at epoch $200$ and
$300$. The maximum training epoch is set to $400$. We set the regularization
parameter, \textit{weight decay} during our training process to $4.0e^{-5}$,
which is used in the Inception model \cite{szegedy2015going}. The weight decay
factor is the same for all the convolution layers in ANTNet. We use the same
default data augmentation module as in ResNet for fair comparison. Random
cropping and horizontal flipping are used for training images and images are
resized or cropped to $224 \times 224$ pixels for ImageNet and $32 \times 32$
pixels for CIFAR100. During test, the trained model is evaluated on center crops.
The same default settings are used in image preprocessing for
evaluation as ResNet \cite{he2016deep}.

\subsection{CIFAR100 Classification}
The CIFAR100 dataset consists of $32\times 32$ RGB images of 100
classes, with $50, 000$ training images and $10, 000$ test images. We consider
the start-of-the-art network architecture MobileNetV2 as our baseline. For fair comparison,
we keep our settings the same as MobileNetV2. The $32\times 32$ images are converted to
$40\times 40$ images with zero-padding by $4$ pixels on each side.
Then, we randomly sample a $32 \times 32$ crop from the $40 \times 40$
image. Horizontal flipping and RGB mean value subtraction are applied as well.
The overall network architecture and the hyperparameters for CIFAR100 are
the same as ANTNet for ImageNet described in Table \ref{tab:antnet} except for different
input and output size (100 classes vs. 1,000 classes) and strides of the first conv2d and
the ANTBlock with $14 \times 14\times 96$ set to 1.

As our purpose is to build resource efficient image classifier on mobile
	platform, we only compare our model with low computational cost models with fewer
	parameters consuming less memory and taking small network width. We consider
	mobile-suitable models, MobileNet and ShuffleNet as our comparison baselines. We
	evaluate the top-1 and top-5 accuracy and compare MAdds and number of parameters
	for benchmark. The performance comparison between baseline models and our ANTNet
	is listed in the table \ref{tab:cifarperf}. It is easy to notice that our ANTNet
	achieve significant improvements over MobileNetV2 and ShuffleNet with fewer
	computational cost and parameter count. Our ANTNet ($g = 2$) achieves $19.6\%$
	computation reduction and $8.3\%$ parameter reduction with $1.5\%$ increase in
	top-1 accuracy. Plus, our ANTNet ($g = 1$) achieves more accuracy improvement
	$1.7\%$ increase of top-1 accuracy with a slightly more computational cost and
	parameter count.

\begin{table}[!htbp]
	\centering
	\resizebox{\columnwidth}{!}{
		\begin{tabular} {|c|c|c|c|c|}
			\hline
			\mr{Network} & Top-1  & Top-5  &\mr{\#Parameters} & \mr{\#MAdds}\\
			&        Accu. & Accu. &    & \\
			\hline
			ShuffleNet (1.5) & $70.0$ & $90.78$ & $2.3M$ & $91.0M$ \\
			MobileNetV2 & $74.2$ & $93.3$ &  $2.4M$ & $91.1M$ \\
			ANTNet (g = 1) & $75.9$ & $94.3$ & $2.7M$ & $91.4M$ \\
			\textbf{ANTNet (g = 2)}  & \bm{$75.7$} & $93.6$ & \bm{$2.2M$} & \bm{$73.2M$} \\
			e-ANTNet & \bm{$76.7$} & $94.1$ & $4.4M$ & $154.9M$ \\
			\hline
		\end{tabular}
	}
	\caption{\label{tab:cifarperf} \footnotesize Performance on CIFAR100. We
			compare ANTNet models with MobileNetV2. Our proposed model ANTNet ($g = 2$) achieves $19.6\%$
			computation reduction and $8.3\%$ parameter reduction with $1.5\%$ increase in
			top-1 accuracy.}
\end{table}

\subsubsection{Optimal Configuration of Channel Attention}
Channel attention in the ANTBlock is a key to improve the feature representation
but the naive combination of channel attention and Depthwise Separable
Convolution does not necessarily yield better performance. \hjk{Table
\ref{tab:cantnet} shows that the naive adaptation of
squeeze-and-excitation \cite{hu2018senet} for MobileNetV2 (Se-MobileNetV2) does
not improve the representation power. To combine channel
attention with depthwise convolutional layers, it needs a more careful
design. We observed that channel attention is effective when the
number of channels is large. Also similar to Rule
for Full Channel Receptive Field (FCRF) \cite{zhang2017clcnet}, we design the ANTBlock
that each output channel of a depthwise convolutional layer has a full channel receptive field
to maximize the representation power.
So, channel attention is inserted between expansion and projection layers in
the ANTBlock as proposed in Fig.\ref{fig:antnetblock} (b).
One additional advantage of this design is that since any output channel of the depthwise
convolutional layer has a FCRF, the projection layer in the ANTBlock can be substituted with
any group convolutional layers ensuring that all output channels of an ANTBlock
have a FCRF. All ANTBlocks ($g=1/g=2$) have FCRF.
}

\begin{table}[!h]
	\centering
	\begin{tabular} {|c|c|c|}
		\hline
		Network  & Top-1 Accu. & Top-5 Accu.\\ \hline
		MobileNetV2 & $74.2$ & $93.3$ \\
		se-MobileNetV2 & $74.1$ & $92.8$ \\
		c-ANTNet &  $73.4$ & $93.3$  \\
		ANTNet-c  &  $74.4$ & $93.5$ \\
		\textbf{ANTNet (proposed)} &  \textbf{75.7} & \textbf{93.6}\\
		\hline
	\end{tabular}
	\caption{\label{tab:cantnet} \footnotesize Different configurations of channel
		attention in the ANTBLock are evaluated on CIFAR100. For projection, all ANTNets
		use group convolution ($g=2$) and MobileNetV2 uses the standard convolution
		($g=1$). All the blocks have similar computational cost and parameters. \hjk{
			Our construction (ANTNet) with channel attention between the depthwise
			convolution layer and projection layer shows the largest improvement (1.5\%). It
			is consistent with our intuition. Note that a naive adaptation of
			Squeeze-and-excitation does \textit{not} improve the performance of
			MobileNetV2. se-MobilenetV2, which has a simple concatenation of a MobileNetV2
			block and a SE-Block, shows degradation compared to MobileNetV2. Even in a
			mobileNetV2 block, channel attention at arbitrary layers such as before the
			expansion layer (c-ANTNet) and after the projection layer (ANTNet-c) are not
			effective.}}
	\vspace{5pt}
\end{table}

\hjk{We compare the accuracy of three different arrangements of channel
attention against the MobileNetV2 and they have similar computational costs and
parameter counts. Our experiments in Table \ref{tab:cantnet}  show that ANTBlock
with channel attention between expansion and projection layers was most
effective (+1.5\% Top-1 Accuracy) whereas all other arrangements do not show a
significant performance boost. The channel attention after the projection layer
(ANTNet-c) has almost the same performance as MobileNetV2 and channel attention
before the expansion layer (c-ANTNet) even reduces the representational power.
This experimental result is consistent with our observation and shows that
channel attention is most effective with a large number of channels and a full
channel receptive field.}

\subsubsection{Reduction Ratio}
Reduction ratios $r_i, \;i = 1, \dots, N$, in Eq. \eqref{eq:cparams} are hyperparameters to adjust the capacity and MAdds/Params. We varied $r_i$ at each ANTBlock and our final model (ANTNet) achieved the better accuracy (see Table \ref{tab:ratio}) with less parameters rather than fixed $r_i$ for all ANTBlocks. We also observed that the last stage of the network shows an interesting tendency towards a saturated state. We found that the setting of reduction ratio for our ANTNet (see Table \ref{tab:antnet}) achieved a good balance between accuracy and complexity and we thus use this setting for all experiments.

\begin{table}[!b]
	\centering
	\caption{\label{tab:ratio}\footnotesize Performance comparison of our ANTNet with the configuration of using different reduction ratios $r_i$ for each ANTBlock and fixed $r_i$ for all ANTBlocks on CIFAR100.}
	{\footnotesize
	\begin{tabular}{|c|c|c|c|c|}
		\hline
		Ratio $r_i$ & Params & MAdds & Top-1 accu & Top-5 accu\\ \hline
		8 & 3.5M & 92.3M & 75.9 & 94.1\\
		16 & 3.0M & 91.7M & 75.2 & 93.9\\
		32 & 2.8M & 91.5M & 75.5 & 93.9\\
		\textbf{Ours (mixed)} & \textbf{2.7M} & \textbf{91.4M} & \textbf{75.9} & \textbf{94.3}\\ \hline
	\end{tabular}
}
\end{table}

\subsubsection{Parameters Learning of e-ANTBlock}
Adaptively weighting different types of ANTBlocks, e-ANTBlock, allows building larger models with higher accuracy. In our experiment, we use $m = 2$ types of ANTBlocks (varying number of groups in convolution) by constructing e-ANTBlock. $j \in \{1, 2\}$ indicates we are using two types of ANTBlocks with group convolution with group $g_1 = 1$ and group $g_2 = 2$. $w_1$ and $w_2$ are the parameters corresponding to their weights of two types of ANTBlocks. We compared the accuracy on CIFAR100 with manually set weight parameters $(w_1, w_2)$, e.g., (0,1), (1,0), (0.5,0.5), etc. If we set $w_1 = 1, w_2 = 0$, or $w_1 = 0, w_2 = 1$, it means we are using only one type of ANTBlock for constructing e-ANTBlock and another one type of ANTBlock is not used. When we set $w_1 = 0.5, w_2 = 0.5$, it means we are using both type of blocks for constructing e-ANTBlock by averaging. The experiment on CIFAR100 with e-ANTNet shows that automatic learning $w_g$ outperforms manually setting (see Table \ref{tab:agg}). The best Top1-Accuracy by fixed weights was 76.2\% whereas learned $(w_1, w_2)$ achieved 76.7\%.

 \begin{table}
 	\centering
 	\caption{\label{tab:agg}\footnotesize Performance of e-ANTNet with e-ANTBlocks by adaptively weighting two types of ANTBlock on CIFAR100 w/o learning parameters $w_1$ and $w_2$.}
 	{\footnotesize
 		\begin{tabular}{|c|c|c|}
 			\hline
 			$w_1$  & $w_2$  & Top-1 accu  \\ \hline
 			0 & 1   & 75.7 \\
 			1 & 0 & 75.9  \\
 			0.5 & 0.5 & 76.2\\
 		    learned & learned & 76.7\\ \hline
 		\end{tabular}
 	}
 \end{table}
\subsection{ImageNet Classification}
The ImageNet 2012 dataset consists of $1.28$ million training images and $50$K
validation images from 1,000 classes. We train our network on the training set
and report top-1 and top-5 accuracy with the corresponding MAdds and the parameters
of models.

Our ANTNet achitecture is shown in Fig \ref{fig:antnetblock}(a) and the details
of layers are listed in the Table \ref{tab:antnet}.  We compare our models with
other low-cost models (e.g., $3.4$ M params, $<300$M MAdds), such as
MobileNetV1, MobileNetV2 ($\alpha=1$), and ShuffleNet (1.5). The comparison of
accuracy and computation budgets is shown in Table \ref{lab:imagenet}.  Our
ANTNet ($g = 2$) achieves consistent improvement over MobileNetV2 by $0.8\%$
Top-1 accuracy and outperforms ShuffleNet (1.5) by $1.3\%$. Compared with the
most resource-efficient model, CondenseNet (G=C=4), our ANTNet performs better
than it with $1.8\%$ accuracy improvement, even with fewer MACCs.
With slight more parameters and MACCs, our ANTNet ($g=1$) can offer $1.2\%$
Top-1 accuracy improvement against MobileNetV2 ($\alpha=1$).
Also, we have a variant of our ANTNet which has comparable
performace as MobileNetV2 with similar MAdds and Params as CondenseNet (G=C=4).
\begin{table*}[ht!]
	\centering
	{ \begin{tabular} {|c|c|c|c|c|}
			\hline
			Model & \#Parameters & \#MAdds & Top-1 Accu. (\%) & Top-5 Accu. (\%)  \\
			\hline
			MobileNetV1 & 4.2M & 575M & 70.6 & 89.5  \\
			SqueezeNext & 3.2M & 708M & 67.5 & 88.2  \\
			ShuffleNet (1.5) & 3.4M & 292M & 71.5 & - \\
			ShuffleNet (x2) & 5.4M & 524M & 73.7 & -   \\
			CondenseNet (G=C=4) & 2.9M & \textbf{274M} & \textbf{71.0} & 90.0 \\
			CondenseNet (G=C=8) & 4.8M & 529M & 73.8 & 91.7  \\
			MobileNetV2 & 3.4M & \textbf{300M} & \textbf{72.0} & 91.0  \\
			MobileNetV2 (1.4) & 6.9M & 585M & 74.7 & 92.5 \\
			NASNet-A & 5.3M & 564M & 74.0 & 91.3 \\
			AmoebaNet-A & 5.1M & 555M & 74.5 & 92.0  \\
			PNASNet & 5.1M & 588M & 74.2 & 91.9 \\
			DARTS & 4.9M & 595M & 73.1 & 91 \\
			\hline
			\textbf{ANTNet (g = 1) (ours)} & 3.7M & 322M & \textbf{73.2} & 91.2  \\
			\textbf{ANTNet (g = 2) (ours)} & 3.2M & \textbf{267M} & \textbf{72.8} & 91.0 \\
			\textbf{e-ANTNet (ours)}  & 5.5M & \textbf{545M} & \textbf{74.2} & 91.6 \\
			\textbf{ANTNet ($\alpha=1.4$) (ours)} & 6.8M & \textbf{598M} & \textbf{75.0} & 92.3 \\
			\hline
	\end{tabular}}
	\caption{\label{lab:imagenet}\footnotesize Performance Results on ImageNet
		Classification. We compare our AntNet models with mobile models. Our proposed model ANTNet ($g=2$) achieves $0.8\%$ absolute Top-1 accuracy improvement over MobileNetV2 with $6\%$ fewer parameters and $11\%$ fewer MAdds. Compare with the lightest model CondenseNet (G=C=4), our model achieves $1.8\%$ absolute Top-1 accuracy with fewer MAdds. To compare with $\sim 600$M \#MAdds, we increase the dimension of features with depth multiplier ($\alpha=1.4$) of our ANTNet, ANTNet ($\alpha=1.4$), and it performs better than all baseline models, $0.3\%$ Top-1 accuracy improvement over MobileNetV2 ($1.4$), $1\%$ Top-1 accuracy improvement over NASNet-A and $2\%$ Top-1 accuracy improvement over DARTS. }
	\vspace{5pt}
\end{table*}

\subsection{Inference on a Mobile Device}
We briefly discussed that MAdds and Params are used to measure the computational
cost and model size.
They are handy to compare models across a variety of implementation and
hardware. But this estimate does not consider memory reads and writes cost,
which can be a crucial factor in a real world scenario.
Since memory access is relatively slower than
computations, the amount of memory access will have a big impact on its real
speed on actual devices. Moreover, both CPUs and GPUs can do caching to speed up
memory reads and writes. Memory coalescing can be very useful for speeding
up memory reads as each thread can read a chunk of memory in one go instead of
doing separate reads. Kernels can also read small amounts of memory into
local or thread group storage of faster access.  It is possible for each thread
to compute multiple outputs instead of only one, allowing it to reuse some of
the input multiple times and thus requiring fewer memory reads overall.
In short, the actual inference speed running on actual devices depends on
hardware architecture and the ways of implementation of each layer.
So the inference speed of models should be tested on actual devices as well.

\begin{table}[!b]
\centering
\resizebox{\columnwidth}{!}{
\begin{tabular} {|c|c|c|c|}
\hline
Model &  MAdds & CoreML Model Size & Latency\\
\hline
MobileNetV2 &  $300M$ & $14.7 M$ & $197.2ms$ \\
ANTNet (g = 1) &  $322M$ & $15.8 M$ & $214.2ms$\\
\textbf{ANTNet (g = 2)}  & $\mathbf{267 M}$ & $\mathbf{13.4M}$ & $\mathbf{157.7ms}$ \\
\hline
\end{tabular}
}
\caption{\label{tab:imageiphone}\footnotesize Latency (inference time) running on an actual device, iPhone 5s. Our proposed model ANTNet (g = 2) achieves $20\%$ faster than MobileNetV2. }
\end{table}

We evaluate the actual inference time of models on a commodity iOS-based
smartphone \textit{iPhone 5s}, which has a 64-bit 1.3 GHz dual-core Apple
Cyclone, Apple A7, Apple M7 motion coprocessor and 1GB LPDDR3 RAM. To run the
inference of models on iPhone5s, we need to convert our trained models to
\textit{CoreML} models, which can be deployed on iOS-based devices using an
Apple machine learning platform. CoreML is optimized for on-device performance
and minimizes memory footprint and power consumption. Although it is only
focused on and optimized on iOS-based platform, it can still be meaningful to
compare the speed of ANTNet relative to other baseline models. The actual
inference time of models on iPhone 5s is available in Table
\ref{tab:imageiphone}. We run each model $10$ times and take out the fastest and
the slowest runs, and then take average of $8$ runs as the final inference time. The table
also provides converted CoreML model file sizes. Table \ref{tab:imageiphone}
shows that our ANTNet $(g=2)$ achieves $20\%$ speedup compared to MobileNetV2
and the improvement of latency is our analysis of MAdds.

\hjk{The CPU inference time on a desktop
machine with a 2.10 GHz 32-core Intel(R) Xeon(R) CPU E5-2620 shows
similar improvement as \textit{iPhone 5s} that our ANTNet ($g=2$) is 8\% faster than
MobileNetV2 (1.11s vs 1.21s).}

\section{Conclusion}
In this paper we proposed the ANTBlock, a novel basic architecture unit designed
to boost the representational capacity of a network by imposing channel-wise
attention and grouped convolution. The capacity of ANTBlock allows designing
resouce-efficient networks. MobileNetV2 can be viewed as a special case of our
network with the removal of channel-wise attention and group convolution.
Extensive experiments demonstrate the effectiveness and efficiency of our ANTNet
which achieves state-of-the-art performance on multiple datasets. In addition,
the experiments on an actual device iPhone 5s show that ANTNet achieves
significant latency improvement on top of state-of-the-art low cost models in
practice. \hwchange{Finally, the improved capacity induced by ANTBlocks shows that
leveraging the interdependency of channels is a promising direction to find more
resource-efficient mobile models by imposing MAdds and parameter constraints.}

\noindent {\bf Acknowledgments.}
We thank Ambrish Tyagi, James M. Rehg and Yadunandana Rao for discussions.
 \label{sec:conc}

\clearpage
{\small
\bibliographystyle{ieee_fullname}
\bibliography{antnetsref}

\begin{thebibliography}{10}\itemsep=-1pt

\bibitem{changpinyo2017power}
Soravit Changpinyo, Mark Sandler, and Andrey Zhmoginov.
\newblock The power of sparsity in convolutional neural networks.
\newblock {\em arXiv preprint arXiv:1702.06257}, 2017.

\bibitem{chollet2017xception}
Fran{\c{c}}ois Chollet.
\newblock Xception: Deep learning with depthwise separable convolutions.
\newblock {\em arXiv preprint}, pages 1610--02357, 2017.

\bibitem{han2015learning}
Song Han, Jeff Pool, John Tran, and William Dally.
\newblock Learning both weights and connections for efficient neural network.
\newblock In {\em Advances in neural information processing systems}, pages
  1135--1143, 2015.

\bibitem{he2015convolutional}
Kaiming He and Jian Sun.
\newblock Convolutional neural networks at constrained time cost.
\newblock In {\em Proceedings of the IEEE conference on computer vision and
  pattern recognition}, pages 5353--5360, 2015.

\bibitem{he2016deep}
Kaiming He, Xiangyu Zhang, Shaoqing Ren, and Jian Sun.
\newblock Deep residual learning for image recognition.
\newblock In {\em Proceedings of the IEEE conference on computer vision and
  pattern recognition}, pages 770--778, 2016.

\bibitem{howard2017mobilenets}
Andrew~G Howard, Menglong Zhu, Bo Chen, Dmitry Kalenichenko, Weijun Wang,
  Tobias Weyand, Marco Andreetto, and Hartwig Adam.
\newblock Mobilenets: Efficient convolutional neural networks for mobile vision
  applications.
\newblock {\em arXiv preprint arXiv:1704.04861}, 2017.

\bibitem{hu2017squeeze}
Jie Hu, Li Shen, and Gang Sun.
\newblock Squeeze-and-excitation networks.
\newblock {\em arXiv preprint arXiv:1709.01507}, 7, 2017.

\bibitem{hu2018senet}
Jie Hu, Li Shen, and Gang Sun.
\newblock Squeeze-and-excitation networks.
\newblock 2018.

\bibitem{Huang_2018_CVPR}
Gao Huang, Shichen Liu, Laurens van~der Maaten, and Kilian~Q. Weinberger.
\newblock Condensenet: An efficient densenet using learned group convolutions.
\newblock In {\em The IEEE Conference on Computer Vision and Pattern
  Recognition (CVPR)}, June 2018.

\bibitem{huang2017densely}
Gao Huang, Zhuang Liu, Laurens Van Der~Maaten, and Kilian~Q Weinberger.
\newblock Densely connected convolutional networks.
\newblock In {\em CVPR}, volume~1, page~3, 2017.

\bibitem{ioannou2017deep}
Yani Ioannou, Duncan Robertson, Roberto Cipolla, Antonio Criminisi, et~al.
\newblock Deep roots: Improving cnn efficiency with hierarchical filter groups.
\newblock 2017.

\bibitem{ioffe2015batch}
Sergey Ioffe and Christian Szegedy.
\newblock Batch normalization: Accelerating deep network training by reducing
  internal covariate shift.
\newblock {\em arXiv preprint arXiv:1502.03167}, 2015.

\bibitem{jin2014flattened}
Jonghoon Jin, Aysegul Dundar, and Eugenio Culurciello.
\newblock Flattened convolutional neural networks for feedforward acceleration.
\newblock {\em arXiv preprint arXiv:1412.5474}, 2014.

\bibitem{krizhevsky2009learning}
Alex Krizhevsky.
\newblock Learning multiple layers of features from tiny images.
\newblock Technical report, Citeseer, 2009.

\bibitem{nips2012_4824}
Alex Krizhevsky, Ilya Sutskever, and Geoffrey~E Hinton.
\newblock Imagenet classification with deep convolutional neural networks.
\newblock In F. Pereira, C.~J.~C. Burges, L. Bottou, and K.~Q. Weinberger,
  editors, {\em Advances in Neural Information Processing Systems 25}, pages
  1097--1105. Curran Associates, Inc., 2012.

\bibitem{krizhevsky2012imagenet}
Alex Krizhevsky, Ilya Sutskever, and Geoffrey~E Hinton.
\newblock Imagenet classification with deep convolutional neural networks.
\newblock In {\em Advances in neural information processing systems}, pages
  1097--1105, 2012.

\bibitem{long2015fully}
Jonathan Long, Evan Shelhamer, and Trevor Darrell.
\newblock Fully convolutional networks for semantic segmentation.
\newblock In {\em Proceedings of the IEEE conference on computer vision and
  pattern recognition}, pages 3431--3440, 2015.

\bibitem{ma2018shufflenet}
Ningning Ma, Xiangyu Zhang, Hai-Tao Zheng, and Jian Sun.
\newblock Shufflenet v2: Practical guidelines for efficient cnn architecture
  design.
\newblock {\em arXiv preprint arXiv:1807.11164}, 2018.

\bibitem{russakovsky2015imagenet}
Olga Russakovsky, Jia Deng, Hao Su, Jonathan Krause, Sanjeev Satheesh, Sean Ma,
  Zhiheng Huang, Andrej Karpathy, Aditya Khosla, Michael Bernstein, et~al.
\newblock Imagenet large scale visual recognition challenge.
\newblock {\em International Journal of Computer Vision}, 115(3):211--252,
  2015.

\bibitem{sandler2018mobilenetv2}
Mark Sandler, Andrew Howard, Menglong Zhu, Andrey Zhmoginov, and Liang-Chieh
  Chen.
\newblock Mobilenetv2: Inverted residuals and linear bottlenecks.
\newblock In {\em Proceedings of the IEEE Conference on Computer Vision and
  Pattern Recognition}, pages 4510--4520, 2018.

\bibitem{sifre2014rigid}
Laurent Sifre and St{\'e}phane Mallat.
\newblock {\em Rigid-motion scattering for image classification}.
\newblock PhD thesis, Citeseer, 2014.

\bibitem{srivastava2015training}
Rupesh~K Srivastava, Klaus Greff, and J{\"u}rgen Schmidhuber.
\newblock Training very deep networks.
\newblock In {\em Advances in neural information processing systems}, pages
  2377--2385, 2015.

\bibitem{szegedy2015going}
Christian Szegedy, Wei Liu, Yangqing Jia, Pierre Sermanet, Scott Reed, Dragomir
  Anguelov, Dumitru Erhan, Vincent Vanhoucke, and Andrew Rabinovich.
\newblock Going deeper with convolutions.
\newblock In {\em Proceedings of the IEEE conference on computer vision and
  pattern recognition}, pages 1--9, 2015.

\bibitem{xie2018interleaved}
Guotian Xie, Jingdong Wang, Ting Zhang, Jianhuang Lai, Richang Hong, and
  Guo-Jun Qi.
\newblock Interleaved structured sparse convolutional neural networks.
\newblock In {\em Proceedings of the IEEE Conference on Computer Vision and
  Pattern Recognition}, pages 8847--8856, 2018.

\bibitem{xie2017aggregated}
Saining Xie, Ross Girshick, Piotr Doll{\'a}r, Zhuowen Tu, and Kaiming He.
\newblock Aggregated residual transformations for deep neural networks.
\newblock In {\em Computer Vision and Pattern Recognition (CVPR), 2017 IEEE
  Conference on}, pages 5987--5995. IEEE, 2017.

\bibitem{zhang2017interleaved}
Ting Zhang, Guo-Jun Qi, Bin Xiao, and Jingdong Wang.
\newblock Interleaved group convolutions.
\newblock In {\em Computer Vision and Pattern Recognition}, 2017.

\bibitem{Zhang_2018_CVPR}
Xiangyu Zhang, Xinyu Zhou, Mengxiao Lin, and Jian Sun.
\newblock Shufflenet: An extremely efficient convolutional neural network for
  mobile devices.
\newblock In {\em The IEEE Conference on Computer Vision and Pattern
  Recognition (CVPR)}, June 2018.

\bibitem{zhang2017clcnet}
Dong-Qing Zhang~et al.
\newblock clcnet: Improving the efficiency of convolutional neural network
  using channel local convolutions.
\newblock {\em arXiv preprint arXiv:1712.06145}, 2017.

\end{thebibliography}
}

\end{document}